# Application of Decision Tree Classifier in Detection of Specific Denial of Service Attacks with Genetic Algorithm Based Feature Selection on NSL-KDD

Deanna M. Wilborne

*Abstract*—Increasing use of communication via the Internet, as well as increasing connectivity speed, has increased the need to more efficiently and quickly detect network intrusion and attack. Possible approaches to increasing the speed and accuracy of detection include decreasing the number of features used by supervised learning methods for intrusion detection, identifying the ideal subset of selected features for identifying intrusions, and focusing on specific attacks for parallelization of intrusion detection. Many proposed network intrusion detection methods use a Genetic Algorithm to reduce the number of features selected and to select the best features for classification. We propose the use of a Genetic Algorithm for the global feature search while using a Decision Tree Classifier for determining the fitness. The goals are to select the minimum required features for each specific attack studied, reduce false positives, and reduce classification time once a classifier is trained.

## I. INTRODUCTION

There are several recent articles that describe supervised learning for intrusion detection systems (IDS) using feature set selection optimized with a Genetic Algorithm [1][2][3]. In the UCI KDD99 Network Intrusion Database, network data is represented as flows between hosts and features include items such as duration, protocol type, source bytes, destination bytes, and etcetera for a total of forty-one (41) features [4]. One goal of Genetic Algorithms (GA) when used in IDS is to find the optimal feature set for allowing maximum positive detection and minimized false positives and false negatives. The optimal feature set is typically a reduced set of total features available [5][6].

To optimize feature selection, genetic algorithms start with an initial randomly selected population of chromosomes and evolve the population through a number of generations using crossover and mutation operators. A fitness function is used to select members of the population that are preferentially chosen for crossover and survival into subsequent generations. After some number of generations or by reaching a specified level of fitness or time elapsed the GA stops, and the most fit member of the population is used as the final and most optimal solution.

Genetic algorithms have a time and space complexity [7]. Time as measured by the total time taken to find an optimal solution and space in the amount of memory required to store the population and other data required by the fitness function. For IDS, a dataset is used by the fitness function to determine how selected features can be used to classify network traffic as an attack or normal [8]. Once an optimal feature set for an attack is found and a classifier is trained this time and space complexity is no longer a component of an IDS.

## II. RELATED WORK

Network utilization is growing in terms of both the amount of data and the number of users. It is important to keep systems and data we access safe and secure [5], [9]–[11]. The literature supports using genetic algorithms as part of the solution to IDS rather than relying on a simple approach based on a decision tree classifier using all or arbitrarily selected features. The literature also supports classification methods that use artificial neural networks or support vector machines [1]–[3][12].

With the increase in the amount of data to be checked for intrusions increasing, the time complexity of IDS has also increased prompting the use of genetic algorithms to reduce the search space complexity by optimizing the feature set used [7]. Time complexity of genetic algorithms can be controlled by the number of generations the GA is allowed to evolve or by the total elapsed time or fitness value preferred. The evolutionary approach can lead to overfitting of the training data and care must be taken to minimize false positives without sacrificing the accuracy of correctly classifying normal or abnormal activity. Fine tuning of fitness parameters can be done by trial and error [13]. Overfitting of data can result in the failure to recognize new or unknown attacks [7]. Each of the aspects discussed above can be fine-tuned to adjust the time complexity of the evolutionary process. These parameters can be thought of as another search space and a GA can be used to search this space.

The related work primarily deals with finding feature sets for the four large classes of attacks provided in the KDD Cup '99 database and derivatives of that database: Probe, User to Root (U2R), Remote to Local (R2L), and Denial of Service (DoS). Each class has several attacks that are members of the class [14]. Classes of attacks and individual attacks can have different relevant features [15].

## III. RESEARCH METHODOLOGY

We propose the use of the NSL-KDD dataset for our experiments [16]. The NSL-KDD dataset has several advantages that include removal of redundant records from the original KDD Cup '99 dataset [4], removal of invalid records, and are smaller allowing the full training and test datasets to be used for our decision tree classifier without the need to partition the datasets into multiple training and test datasets [14]. The number of training and test samples for the NSL-

TABLE I
NSL-KDD DATASET

| Class of Attack | Training Samples | Test Samples |
|---|---|---|
| Normal | 67,343 | 9,711 |
| DoS | 45,927 | 5,741 |
| Probe | 11,656 | 1,106 |
| R2L | 995 | 2,199 |
| U2R | 52 | 37 |
| Unknown | 0 | 3,750 |
| Total | 125,973 | 22,544 |

KDD dataset are shown in Table I by Class of Attack.

For our experiments we will focus on the specific Denial of Service attacks available in the NSL-KDD dataset. The number of training and test samples are shown in Table II.

TABLE II
DOS ATTACKS

| Class of Attack | Training Samples | Test Samples |
|---|---|---|
| Neptune | 41,214 | 4,657 |
| Smurf | 2,646 | 665 |
| Back | 956 | 359 |
| Teardrop | 892 | 12 |
| PoD | 201 | 41 |
| Land | 18 | 7 |
| Total | 45,927 | 5,741 |

In order to allow our research and results to be easily reproducible we propose the use of Python and Scikit-Learn's Decision Tree Classifier [17]. Python is an interpreted language that is available for many operating systems and allows for rapid prototyping of ideas [18]. Our source code and pre-processed data are available by sending an e-mail to mw1313@mynsu.nova.edu. The original NSL-KDD data is available from [16].

To use Scikit-learn's Decision Tree Classifier, we preprocessed the NSL-KDD datasets. We converted all symbolic values, for example, protocol types TCP, UDP, ICMP, service types such as FTP, HTTP, etc., to numbers.

For our Genetic Algorithm each chromosome, or member of the population, will have forty-one (41) binary genes. Each gene corresponds to one of the features in the dataset. A gene value of zero (0) indicates the feature will not be included in training and testing of the classifier. A gene value of one (1) indicates the feature will be included.

Our pseudo code for our Genetic Algorithm and Fitness function, that uses the Decision Tree Classifier, and a flow chart (Figure 1) are shown below:

```
Procedure GA
    Begin
        Create initial population;
        computeFitness(population);
        While termination condition is not true do:
            Begin
                Select parents from population;
                Perform crossover operation and create children;
                Perform mutation operation on children;
                computeFitness(children);
                Add children to population;
                Sort population by fitness;
                Retain top population;
            End
    End

Function computeFitness(population):
    Begin
        For each member of the population:
            Select training data from features of member;
            Train Decision Tree classifier with training data;
            Select test data from features of member;
            Validate tree classifier with selected test data;
            Add true positive, false positive, true negative,
                false negative, and f-measure to member;
        Next member
    End
```

For classifier validation, the f-measure is computed from the true positive (TP), false negative (FN), and false positive (FP) values collected during validation. The f-measure is calculated using equations 1, 2 and 3 [19]:

$$precision = \frac{TP}{TP + FP} \quad (1)$$

$$recall = \frac{TP}{TP + FN} \quad (2)$$

$$f_{measure} = \frac{2 \cdot precision \cdot recall}{precision + recall} \quad (3)$$

Once the f-measure for a member of the population is calculated the fitness of each member of the population using equation (4):

$$fitness = 1 - f_{measure} \qquad (4)$$

Lower fitness values are better. Zero (0) represents perfect fitness. Members of the population are first sorted by fitness then by selected feature count. The goal is to get the best fitness with the least number of features.

In addition to the formulas above used to drive the search of the Genetic Algorithm, the following equations for accuracy and specificity are also calculated and stored with the results for comparison purposes [20]:

$$accuracy = \frac{TP + TN}{TP + FP + FN + TN} \qquad (5)$$

$$specificity = \frac{TN}{FP + TN} \qquad (6)$$

For each run of the training data all attacks that are not the target of classification are treated as normal data. This is done to minimize the influence other attacks have on the classification of the target attack. This can also be beneficial to a parallel ensemble classifier where multiple dedicated classifiers process the same data stream in parallel with each classifier searching for a particular attack, ignoring other attack types along with normal data.

## IV. EXPERIMENTAL RESULTS

Table III shows the result of classification of all Denial-of-Service attacks. The decision tree was built using the split criterion of entropy or information gain. Total features were reduced by the Genetic Algorithm from forty-one (41) to eleven (11). These results were obtained using a population of one hundred (100) chromosomes and eighty (80) generations of processing. Mutation probability was set to 0.024. The selected features are listed in Appendix A.

Table IV shows the result of classification of all Denial of Service attacks using the Gini (information impurity) split criterion [17]. The same population size, generations and mutation were used as for the results in Table III. Appendix B lists the selected features from Table IV.

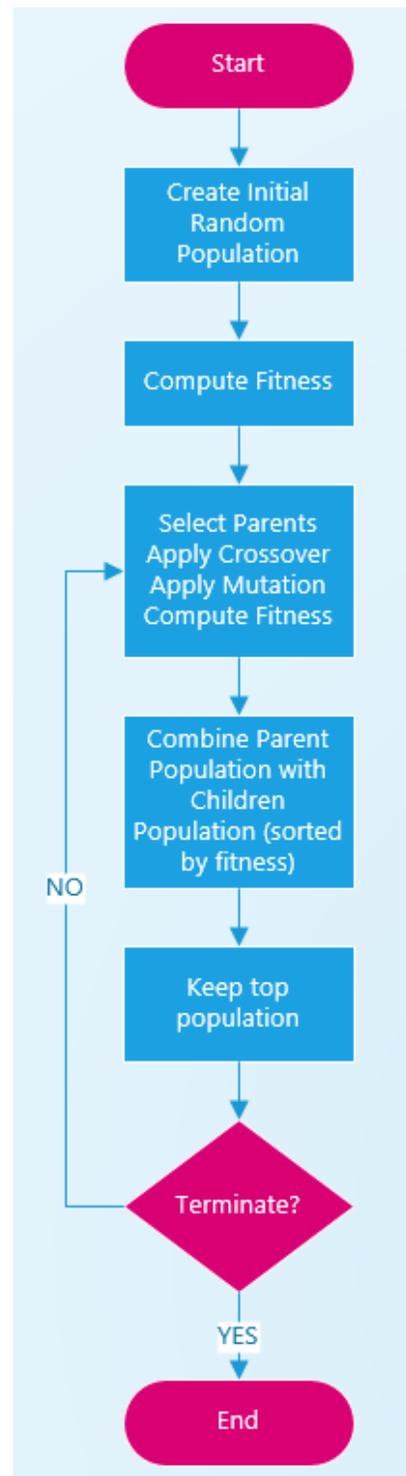

Fig. 1. Flow Chart of Genetic Algorithm

TABLE III
DoS (ALL ATTACKS) USING ENTROPY (INFORMATION GAIN)

| Features | Attack Type | True Positive | False Negative | False Positive | True Negative | Total | Accuracy | Precision | Sensitivity / Recall | F-Measure | Specificity |
|---|---|---|---|---|---|---|---|---|---|---|---|
| 11 | DoS | 5718 | 23 | 57 | 16746 | 22544 | 99.65% | 99.01% | 99.60% | 99.31% | 99.66% |

TABLE IV
DoS (ALL ATTACKS) USING GINI (INFORMATION IMPURITY)

| Features | Attack Type | True Positive | False Negative | False Positive | True Negative | Total | Accuracy | Precision | Sensitivity / Recall | F-Measure | Specificity |
|---|---|---|---|---|---|---|---|---|---|---|---|
| 11 | DoS | 5716 | 25 | 54 | 16749 | 22544 | 99.65% | 99.06% | 99.56% | 99.31% | 99.68% |

TABLE V
DoS ATTACKS USING AN ARTIFICIAL NEURAL NETWORK CLASSIFIER [21]

| Features | Attack Type | Detection Rate | False Positive | False Negative |
|---|---|---|---|---|
| 41 | back | 100.00% | 0.00% | 0.00% |
| 41 | land | 100.00% | 0.00% | 0.00% |
| 41 | neptune | 100.00% | 0.00% | 0.00% |
| 41 | pod | 99.01% | 0.99% | 0.00% |
| 41 | smurf | 99.01% | 0.99% | 0.00% |
| 41 | teardrop | 79.00% | 15.00% | 6.00% |
|  | Average | 96.17% | 2.83% | 1.00% |

TABLE VI
DoS ATTACKS USING DECISION TREE CLASSIFIER

| Features | Attack Type | Detection Rate | False Positive | False Negative | Split Criterion |
|---|---|---|---|---|---|
| 11 | DoS | 98.88% | 0.99% | 0.14% | entropy |
| 11 | DoS | 99.24% | 0.32% | 0.44% | gini |

The results for the Gini split criterion are slightly better compared to entropy. However, both of the results in Table IV are better than those obtained by Ahmad *et al.* [21] using an artificial neural network and all forty-one (41) features. Table V, which is reconstructed from the data in their paper, shows the results by Ahmad *et al*. The equation for the Detection Rate is show in (7).

$$detection_{rate} = 100\% - FP\% - FN\% \qquad (7)$$

The detection rate for the entropy and Gini based decision tree classifiers, for the entire class of Denial-of-Service attacks, are shown in Table VI.

The detection rate for the DoS class of attacks is better than the average detection rate of the artificial neural network. It also uses thirty (30) less features compared to the artificial neural network that uses forty-one (41).

The results in Tables III-VI establish a baseline for comparison of feature reduction and accuracy of classification for each individual attack in the Denial-of-Service class. The values of *f-measure* and **detection rate** will be used to show how classification for each attack is better or worse than the results established.

Table VII shows the results of using entropy and running multiple experiments on each of the six (6) Denial of Service attacks: *back, land, neptune, pod, smurf,* and *teardrop*. The feature count is greatly reduced and accuracy of classification for the attacks *back* and *land* are as good as those obtained in the paper by Ahmad *et al* [21]. Detection rates for *smurf* and *teardrop* are better while *neptune* is slightly lower and *pod* is several percent lower. On average, however, the results are better than using all forty-one (41) features of the artificial neural network.

TABLE VII
DoS ATTACKS USING ENTROPY DECISION TREE

| Features | Attack Type | Detection Rate | False Positive | False Negative |
|---|---|---|---|---|
| 2 | back | 100.00% | 0.00% | 0.00% |
| 1 | land | 100.00% | 0.00% | 0.00% |
| 8 | neptune | 99.80% | 0.07% | 0.13% |
| 2 | pod | 95.06% | 0.07% | 4.88% |
| 2 | smurf | 100.00% | 0.00% | 0.00% |
| 2 | teardrop | 99.84% | 0.16% | 0.00% |
|  | Average | 99.11% | 0.05% | 0.83% |

Table VIII shows the complete results of the experiments using entropy as the split criterion for the decision tree classifier along with the genetic algorithm for feature selection. Appendix C lists the features identified for each attack.

Experiments were also run for *neptune, pod*, and *teardrop* using the gini split criterion for the decision tree classifier to see if better results could be obtained. Slightly better results were obtained for *pod* and *teardrop* and are shown in Table IX. However, the features doubled from two (2) to four (4). The features selected are listed in Appendix D. Better results were not obtained for the *neptune* attack.

TABLE IX
DoS ATTACKS USING GINI DECISION TREE

| Features | Attack Type | Detection Rate | False Positive | False Negative |
|---|---|---|---|---|
| 4 | pod | 99.93% | 0.07% | 0.00% |
| 4 | teardrop | 99.84% | 0.16% | 0.00% |

Table X contains the details collected on the pod and teardrop attacks using the split criterion of gini.

V. CONCLUSIONS

TABLE VIII
DOS (INDIVIDUAL ATTACKS) USING ENTROPY (INFORMATION GAIN)

| Features | Attack Type | True Positive | False Negative | False Positive | True Negative | Total | Accuracy | Precision | Sensitivity / Recall | F-Measure | Specificity |
|---|---|---|---|---|---|---|---|---|---|---|---|
| 2 | back | 359 | 0 | 0 | 22185 | 22544 | 100.00% | 100.00% | 100.00% | 100.00% | 100.00% |
| 1 | land | 7 | 0 | 0 | 22537 | 22544 | 100.00% | 100.00% | 100.00% | 100.00% | 100.00% |
| 8 | neptune | 4651 | 6 | 13 | 17874 | 22544 | 99.92% | 99.72% | 99.87% | 99.80% | 99.93% |
| 2 | pod | 39 | 2 | 15 | 22488 | 22544 | 99.92% | 72.22% | 95.12% | 82.11% | 99.93% |
| 2 | smurf | 665 | 0 | 0 | 21879 | 22544 | 100.00% | 100.00% | 100.00% | 100.00% | 100.00% |
| 2 | teardrop | 12 | 0 | 37 | 22495 | 22544 | 99.84% | 24.49% | 100.00% | 39.34% | 99.84% |
| | Total | 5733 | 8 | 65 | | Average | 99.95% | 82.74% | 99.17% | 86.87% | 99.95% |

TABLE X
DOS (INDIVIDUAL ATTACKS) USING GINI (INFORMATION IMPURITY)

| Features | Attack Type | True Positive | False Negative | False Positive | True Negative | Total | Accuracy | Precision | Recall | F-Measure | Specificity |
|---|---|---|---|---|---|---|---|---|---|---|---|
| 4 | pod | 41 | 0 | 16 | 22487 | 22544 | 99.93% | 71.93% | 100.00% | 83.67% | 99.93% |
| 4 | teardrop | 12 | 0 | 36 | 22496 | 22544 | 99.84% | 25.00% | 100.00% | 40.00% | 99.84% |

Reduction of features for classification by attack does result in better classification results compared with classifiers that use more features or cover more than one attack. Features that aren't necessary for classification of a specific attack also adds complexity to the classifier. Building an ensemble of parallel classifiers that each focus on a specific attack ignoring other attacks may be one way of addressing the concern of increasing network traffic and increasing data speeds.

It is also evident that the split criterion can affect the accuracy of classification. Four (4) denial of service attacks worked best with entropy as the split criterion. Two (2) attacks worked better with gini (information impurity) as the split criterion.

## VI. FUTURE WORK

Our future work will include a detailed analysis of each of the attacks in the other classes, U2R, R2L and Probe following the approach we used here for Denial-of-Service attacks. Additionally, the genetic algorithm can be modified to also explore the tree depth with the goal of generating a less complex decision tree. The depth can be dynamically explored by defining a minimal acceptable level of classification accuracy or fitness. A comparison of artificial neural networks with feature selection driven by a genetic algorithm is another area to explore to determine if artificial neural networks with reduced feature sets work better than decision trees or if a combination of neural networks and decision trees create a better ensemble classifier. Like depth exploration with decision trees, the number of hidden layers and nodes in the neural network can be explored using a genetic algorithm.

One risk in this study is that the f-measure used balances precision and recall [20], in other words, the beta value is one (1). It may be of interest to see how classification can be improved by changing the weighting of precision and recall by setting a beta value of 0.5 and 2.

## VII. APPENDIX

*A. Table III features*

'proto_type', 'svc_num', 'dst_bytes', 'land', 'wrong_fragment', 'logged_in', 'count', 'rerror_rate', 'same_srv_rate', 'diff_srv_rate', 'dst_host_srv_rerror_rate'

*B. Table IV Features*

'proto_type', 'svc_num', 'dst_bytes', 'land', 'wrong_fragment', 'count', 'rerror_rate', 'same_srv_rate', 'diff_srv_rate', 'srv_diff_host_rate', 'dst_host_rerror_rate'

*C. Table VIII Features by attack with entropy split criterion*

*back*: 'dst_bytes', 'dst_host_srv_diff_host_rate'

*land*: 'land'

*neptune*: 'duration', 'flag_num', 'land', 'rerror_rate', 'srv_rerror_rate', 'diff_srv_rate', 'dst_host_srv_diff_host_rate', 'dst_host_srv_rerror_rate'

*pod*: 'src_bytes', 'dst_host_diff_srv_rate'

*smurf*: 'proto_type', 'src_bytes'

*teardrop*: 'proto_type', 'wrong_fragment'

*D. Table X Features using gini split criterion*

*pod*: 'proto_type', 'src_bytes', 'wrong_fragment', 'dst_host_diff_srv_rate'

*teardrop*: 'proto_type', 'src_bytes', 'dst_host_srv_count', 'dst_host_diff_srv_rate'